\documentclass[conference]{IEEEtran}
\usepackage{amsmath,graphicx}
\usepackage{booktabs}
\usepackage{multirow}
\usepackage{diagbox}
\usepackage{amssymb}
\usepackage{amsmath}
\usepackage{array}
\usepackage{hyperref}
\usepackage{marvosym}

\IEEEoverridecommandlockouts

\ifCLASSINFOpdf
\else
\fi
\begin{document}
\title{GaitMA: Pose-guided Multi-modal Feature Fusion for Gait Recognition}

\author{
    \IEEEauthorblockN{Fanxu Min, Shaoxiang Guo, Hao Fan\textsuperscript{\Letter}, Junyu Dong\textsuperscript{\Letter}}
    \IEEEauthorblockA{Faculty of Information Science and Engineering\\
    Ocean University of China\\
    Qingdao, China\\
    \{minfanxu, guoshaoxiang\}@stu.ouc.edu.cn\\
    \{dongjunyu, fanhao\}@ouc.edu.cn}
    \thanks{\Letter~Corresponding author.}
}


%Telephone: (800) 555--1212, Fax: (888) 555--1212}


\maketitle

\begin{abstract}
    Gait recognition is a biometric technology that recognizes the identity of humans through their walking patterns.
    Existing appearance-based methods utilize CNN or Transformer to extract spatial and temporal features from silhouettes, while model-based methods employ GCN to focus on the special topological structure of skeleton points. However, the quality of silhouettes is limited by complex occlusions, and skeletons lack dense semantic features of the human body.
    To tackle these problems, we propose a novel gait recognition framework, dubbed Gait Multi-model Aggregation Network (GaitMA), which effectively combines two modalities to obtain a more robust and comprehensive gait representation for recognition. 
    First, skeletons are represented by joint/limb-based heatmaps, and features from silhouettes and skeletons are respectively extracted using two CNN-based feature extractors. 
    Second, a co-attention alignment module is proposed to align the features by element-wise attention. 
    Finally, we propose a mutual learning module, which achieves feature fusion through cross-attention, Wasserstein loss is further introduced to ensure the effective fusion of two modalities. Extensive experimental results demonstrate the superiority of our model on Gait3D, OU-MVLP, and CASIA-B.
\end{abstract}

%

% \fi
\IEEEpeerreviewmaketitle

\section{Introduction}
    Gait recognition has recently gained widespread interest as a biometric technology that recognizes people by their walking patterns. 
    Unlike other biometrics like face, fingerprint, and iris, gait can be captured from a distance in uncontrolled settings without the cooperation of individuals.
    However, this challenging technique still faces many difficulties, including complex backgrounds, severe occlusion, unpredictable illumination, arbitrary viewpoints, and diverse clothing changes. 
    The appearance-based methods mainly extract temporal and spatial features from silhouettes by 2D/3D CNN, Transformer, RNN, and LSTM \cite{1-gaitset, 2-gaitpart, 3-cstl}. They focus on extracting features from the whole gait sequence or adjacent frames, this makes them perform poorly when facing lower-quality silhouettes. 
    The model-based methods \cite{4-PoseGait, 5-GaitGraph, 6-gaitgragh2, 7-gaittr, 8-gpgait} mostly take clear and robust skeletons as the input, skeletons in a video are mainly represented as a sequence of joint coordinates which are extracted by pose estimators \cite{11-HRNet}. 
    Benefiting from the rapid development in pose estimation and the application of Graph Convolutional Network (GCN) \cite{12-GCN}, recent model-based methods could even show competitive results compared to appearance-based methods.
    
    \begin{figure}[t]
        \begin{center}
    	\includegraphics[width=1.0\linewidth]{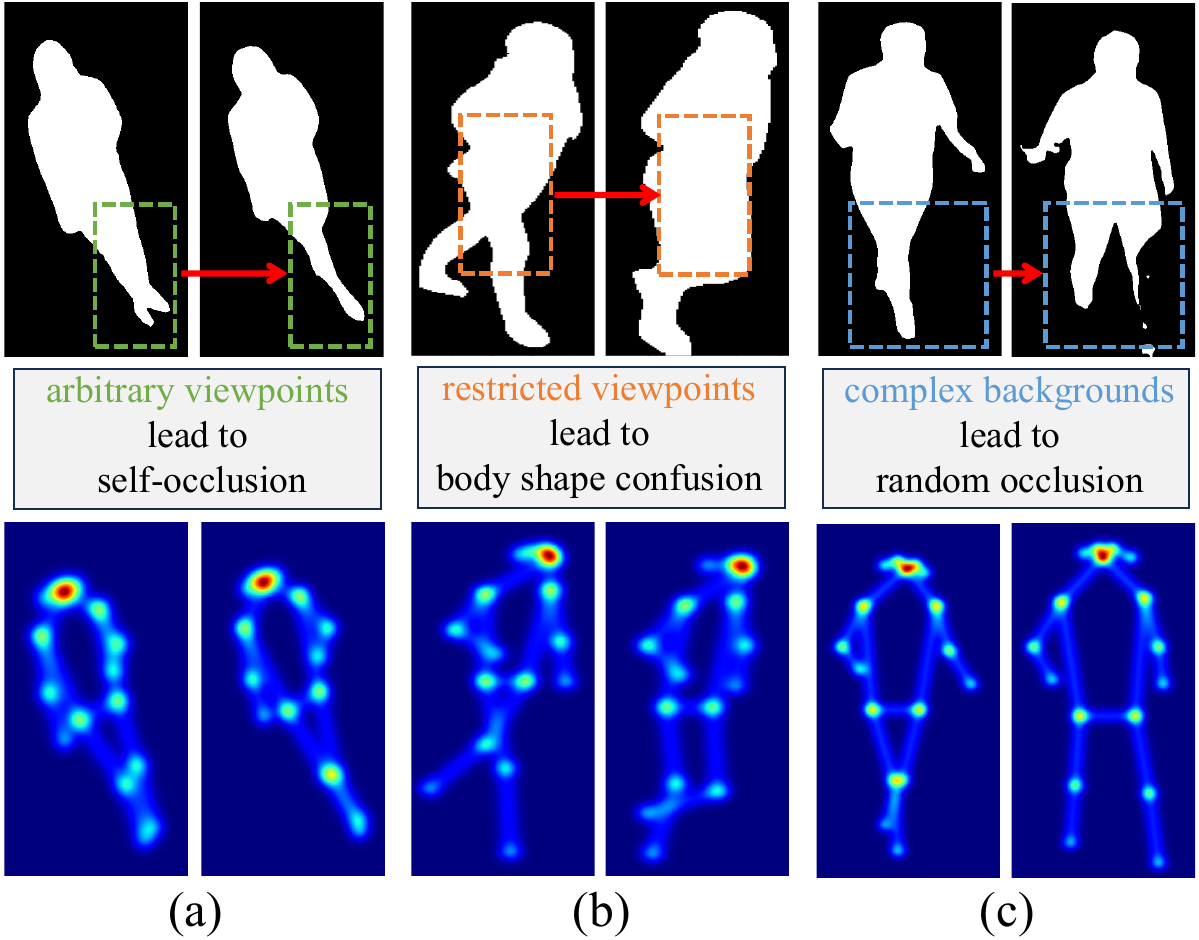}
        \end{center}
        \vspace{-4.0 mm}
        \caption{A brief visualization of our motivation. Skeleton can effectively complement missing gait features in silhouette across various challenging scenarios.}
        \label{fig1}
        % \vspace{-5.0 mm}
    \end{figure} 
    
    \begin{figure*}[ht]
    	\begin{center}
    		\includegraphics[width=1.0\linewidth]{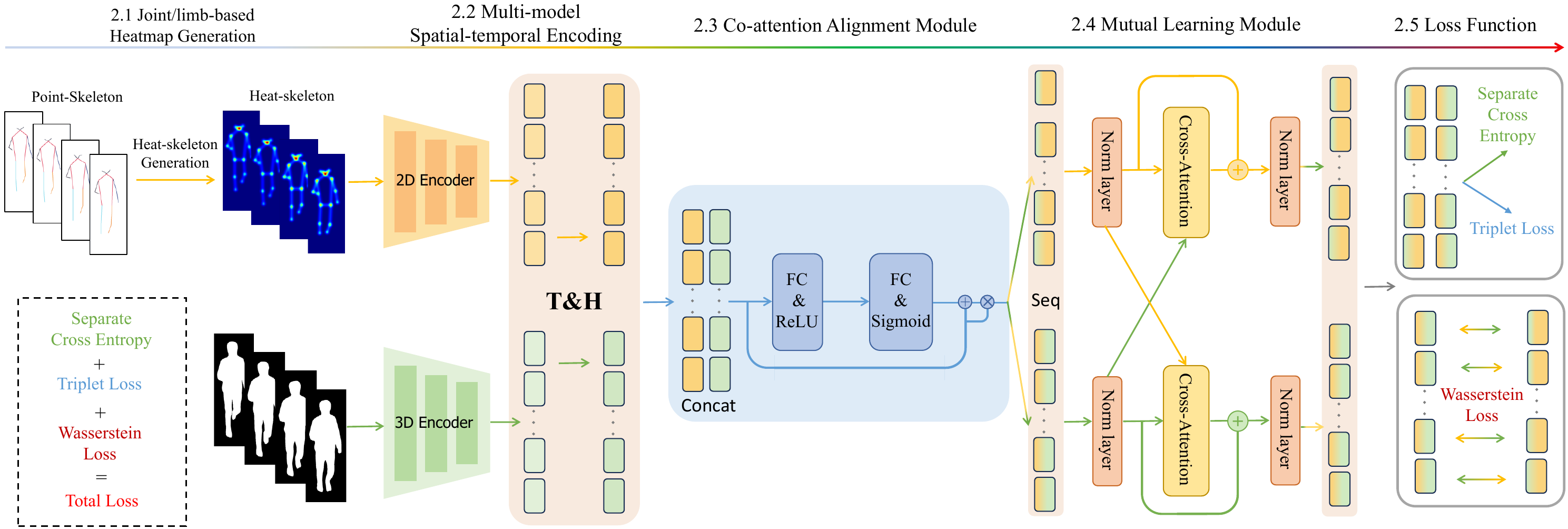}
    	\end{center}
    	% \vspace{-3.5 mm}
    	\caption{An overview of the proposed framework GaitMA for gait recognition. T\&H represents the horizontal mapping and temporal aggregation. Concat and Seq denote the features concatenate and separate, respectively. }
            \label{fig2}
            % \vspace{-2.5 mm}
    \end{figure*}  

    However, modality aggregation in gait recognition is rarely discussed \cite{13-survey-deepgait}. 
    First, as shown in Fig.~\ref{fig1}(a), due to arbitrary viewpoints, the left leg is missing in motion due to self-occlusion, silhouettes can not provide complete gait information in this case, but skeletons give a clear representation of current motion state. 
    Second, due to the problem of self-occlusion in motion, the shape of the human body changes considerably, and it is difficult to distinguish between the torso and the limbs, as shown in Fig.~\ref{fig1}(b), skeletons can guide the posture of the human body to obtain a more robust gait representation. 
    Finally, silhouettes are easily obscured by complex backgrounds and lose shape information, as shown in Fig.~\ref{fig1}(c), skeletons can complement missing gait features in silhouettes. 
    It can be observed that the silhouette retains the external body shape information and omits some body-structure clues, and the skeleton preserves the internal body structure information. The two data modalities are complementary to each other, but they may not correspond, containing mismatched redundancy and interference information.
    Therefore, how to better fuse the silhouette and skeleton is a challenging problem, which significantly influences the performance of obtaining a comprehensive representation of gait. 

    To achieve this goal, we propose a novel gait recognition modality fusion framework, named GaitMA, which effectively combines two modalities to obtain a more robust and comprehensive gait representation for recognition. 
    First, we obtain joint/limb-based heatmaps by computing the Gaussian distribution of skeletal points to enhance the robustness and interoperability of the skeleton \cite{14-revisiting, 15-Heatmap-hand}. This reduces the modality differences between the skeleton and the silhouette. Subsequently, we built a novel asymmetric CNN-based dual-branch architecture to individually extract spatial-temporal gait features from each modality.
    Second, to effectively integrate the two modalities and fully utilize their information, the proposed co-attention alignment module is introduced to mitigate feature redundancy and interference. It achieves alignment by calculating feature attention between elements, thereby bringing the feature distributions of the two modalities closer in the feature space.
    Finally, the mutual learning module is proposed to facilitate the interaction between the two modalities. This module effectively enriches discrete skeleton representations and complements the semantic information of silhouette images. Additionally, Wasserstein loss \cite{21-wasserstein-loss} is introduced to ensure comprehensive mutual learning of features between the two modalities.

    The main contributions of the proposed method are summarized as follows:
    (1) We propose a novel gait recognition modality fusion framework called GaitMA, which utilizes a more comprehensive gait representation constructed from both silhouettes and skeletons represented by joint/limb-based heatmaps to achieve better recognition performance.
    (2) A co-attention alignment module is proposed to improve the efficiency and effectiveness of feature interaction.
    (3) We propose a mutual learning module for feature fusion and introduce Wasserstein loss to ensure effective fusion of the two modalities.
    Experimental results demonstrate that our method achieves superior performance on three dominant datasets, it obtains an average Rank-1 accuracy of 66.1\% on Gait3D, 95.9\% on CASIAB, and 91.2\% on OU-MVLP, respectively. 

\section{METHOD}
    In this section, we will describe the specific details of the model implementation. As shown in Fig.~\ref{fig2}, GaitMA can be divided into five parts: joint/limb-based heatmap generation,  multi-model spatial-temporal encoding, co-attention alignment module, mutual learning module, loss function.

\subsection{Joint/limb-based Heatmap Generation}
    GCN is operated on an irregular graph of skeletons \cite{5-GaitGraph,6-gaitgragh2}, which makes it difficult to fuse with other modalities usually represented on regular grids. we represent each frame of skeleton points as a joint/limb-based heatmap to improve the effectiveness of modality combination \cite{14-revisiting, 15-Heatmap-hand}. By creating Gaussian heatmaps centered at each skeleton point using coordinate triplets $\left( x_{k}, y_{k}, c_{k} \right)$, we obtain the joint-based heatmap $\cal J$ with dimensions of $K\times H\times W$, where K represents the number of joints, and H and W denote the height and width of the frame. The formulation is expressed as,
    \begin{eqnarray}
        {\cal J}_{k i j}&=&e^{-\frac{(i-x_{k})^{2}+(j-y_{k})^{2}}{2*\sigma^{2}}}\ast c_{k}.\label{1}
    \end{eqnarray}
    The parameter $\sigma$ regulates the variance of the Gaussian maps, while $\left( x_{k}, y_{k} \right)$ represents the spatial location of the $k$-th joint, and $c_{k}$ represents the corresponding confidence score. We can also create the limb-based heatmap $\cal L$:
    \begin{eqnarray}
        {\cal L}_{k i j}\,&=&\,e^{-\frac{D((i,j),s e g[a_{k},b_{k}])^{2}}{2\ast\sigma^{2}}}\ast\mathrm{min}\big({\cal C}_{a_{k}}\,,\,{\cal C}_{b_{k}}\big).\label{2}
    \end{eqnarray}
    The limb indexed as k connects two joints, $a_{k}$ and $b_{k}$. The function ${\cal D}$ calculates the distance from the point $\left( i, j \right)$ to the segment $\left[\left(x_{a_{k}},y_{a_{k}}\right),\left(x_{b_{k}},y_{b_{k}}\right)\right]$. Finally, the joint/limb-based heatmap is derived by stacking all the heatmaps for each frame along the K dimension.
\subsection{Multi-model Spatial-temporal Encoding}
    To enhance the efficiency of spatial-temporal feature extraction from gait information while minimizing model size, we introduce an innovative asymmetric CNN-based architecture\cite{17-ResNet} with a dual-branch structure. Opting for a higher resolution of 128x88 in Silhouettes allows for the capture of finer details, whereas a joint/limb-based 64x44 heatmap offers comprehensive spatial shape information with reduced model complexity. The silhouette branch employs a dense 3D-CNN to extract detailed, high-dimensional spatio-temporal features. Concurrently, the skeleton branch supplements feature absent in the silhouette representation, utilizing a streamlined 2D-CNN for spatial feature extraction\cite{9-BiFusion, 10-MMGaitFormer}.
    
    This asymmetric approach effectively consolidates robust features from both modalities and efficiently trims the model's parameter count. The silhouette features $Y_{sil}$ and the skeleton features $Y_{ske}$ are extracted from the silhouette feature extractor and skeleton feature extractor, respectively. 
    After that we introduce horizontal mapping \cite{19-horizontal} and temporal aggregation operations to generate feature representations.
\subsection{Co-attention Alignment Module}
    Silhouette and skeleton features, inherently distinct modalities, often contain mismatched, redundant, and noisy information, impeding detailed inter-modal interactions. Prior research\cite{9-BiFusion, 10-MMGaitFormer} frequently overlooks this complexity, resorting to basic summation, concatenation, or neglecting information redundancy and noise. Addressing this issue, our proposed Co-attention Attention Model (CAM) leverages a self-attention mechanism to align the feature distributions of these two modalities more closely\cite{18-transfomer}. This alignment not only facilitates inter-feature interaction but also enhances the overall efficiency of model fitting.
    
    As illustrated in Figure.~\ref{fig2}, the input $Y_{m}$ is obtained by channel-wise concatenating $Y_{sil}$ and $Y_{ske}$, two fully-connected layers are designed to reduce the number of parameters and achieve the information bottleneck effect. The overall formulation can be expressed as:
    \begin{eqnarray}
        Y_{score} &=& \sigma\left ( \tau\left ( \omega_{1}Y_{m} + b_{1} \right ) \omega _{2} + b_{2} \right),\label{3}
        \\
        Y_{align} &=& Y_{score} \otimes Y_{m} + Y_{m},\label{4}
    \end{eqnarray}   
   $\omega_{1}$, $\omega_{2}$, $b_{1}$, and $b_{2}$ represent the weights and biases of two fully-connected layers, respectively. The symbol $\tau$ denotes the ReLU activation function, while $\sigma$ represents the Sigmoid function. $\otimes$ denotes the element-wise multiplication.
\subsection{Mutual Learning Module}
    To optimize the utilization of features from both modalities, we introduce the mutual learning module (MLM), leveraging a cross-attention mechanism for the comprehensive fusion of these modal features\cite{18-transfomer}. While the CAM facilitates interaction between modal features, primarily aiming to harmonize their distributional variances, our MLM extends beyond this by ensuring a thorough integration. We employ a symmetric dual-branch structure, allowing each modality to focus on its intrinsic information while concurrently enriching the other. This approach not only enhances the discrete skeleton representation but also augments the semantic content of the silhouette images, achieving a balanced and in-depth feature interaction between the modalities. 
    
    The detailed process is shown in Figure.~\ref{fig2}. Take one side for example, assuming that $Y_{1}$ and $Y_{2}$ are the corresponding feature representations of the two modalities, $Y_{1}^{'}$ is the output after mutual learning.
    The formulation is expressed as: 
    \begin{eqnarray}
        Y_{1}^{'}  &=& \Phi\left (\Theta\left(Y_{1}Y_{2}^{T} / \sqrt{d} \right )Y_{2} + Y_{1}  \right ).
    \end{eqnarray}
    $\Phi$ is the layer normalization and $\Theta$ denotes the Softmax function, hyperparameter d denotes the scale factor.

\begin{table*}[ht]
    \footnotesize
    \centering
    \caption{Quantitative comparison of gait recognition methods across three authoritative datasets, involving OUMVLP, GREW, and Gait3D. The best performances are in \textbf{blod}, the second best methods are  \underline{underlined}.}        \label{tab1}
    % \vspace{1mm}
    \begin{tabular}{c|c|ccccccccc}
        \toprule
        \multirow{4}{*}{Modality}  & \multirow{4}{*}{Method} & \multicolumn{9}{c}{Testing Datasets}                                                                 \\
        \cline{3-11}
                                   &                         & \multicolumn{4}{c}{\multirow{2}{*}{Gait3D}} & \multirow{2}{*}{OU-MVLP} & \multicolumn{4}{c}{CASIA-B} \\
        \cline{8-11}                           
                                   &                         & \multicolumn{4}{c}{}                        &                          & NM    & BG    & CL   & Mean \\
        \cline{3-11}
                                   &                         & Rank-1     & Rank-5     & mAP     & mINP    & Rank-1                   & \multicolumn{4}{c}{Rank-1}  \\
        \hline
        \multirow{4}{*}{Sihouette} & GaitSet(AAAI19)\cite{1-gaitset}                 & 36.7             & \underline{59.3} & \underline{30.0} & \underline{17.3} & 87.1                     & 95.0      & 87.2      & 70.4     & 84.2     \\
                                   & GaitPart(CVPR20)\cite{2-gaitpart}                & 28.2             & 47.6             & 21.6             & 12.4             & 88.5                     & 96.2      & 91.5      & 78.7     & 88.8     \\
                                   & GaitGL(ICCV21)\cite{22-GaitGL}                  & 29.7             & 48.5             & 22.3             & 13.6             & 89.7                     & 97.7      & 94.5      & 83.6     & 91.8     \\
                                   & GaitBase(CVPR23)\cite{16-opengait}                & \underline{64.6} & -                & -                & -                & \underline{90.8}         & 97.6      & 94.0      & 77.4     & 89.7     \\
        \hline
        \multirow{4}{*}{Skeleton}  & GaitGraph(ICIP21)\cite{5-GaitGraph}                & 8.6              & -             & -             & -       & 4.2                      & 86.4      & 76.5      & 65.2     & 76.0     \\
                                   & GaitGraph2(CVPRW22)\cite{6-gaitgragh2}              & 11.2             & -             & -             & -       & 70.6                     & 80.3      & 71.4      & 63.8     & 71.8     \\
                                   & GaitTR(ES23)\cite{7-gaittr}                  & 7.2              & -             & -             & -       & 39.7                     & 94.7      & 89.4      & 86.7     & 90.2     \\
                                   & GPGait(ICCV23)\cite{8-gpgait}                  & 22.4             & -             & -             & -       & 59.1                     & 93.6      & 80.2      & 69.3     & 81.0     \\
        \hline
        \multirow{2}{*}{Fusion}    & BiFusion(MTA23)\cite{9-BiFusion}                & -                & -             & -             & -       & 89.9                     & \textbf{98.7}    & \underline{96.0} & 92.1          & 95.6          \\
                                   & MMGaitFormer(CVPR23)\cite{10-MMGaitFormer}            & -                & -             & -             & -       & 90.1                     & \underline{98.4} & \underline{96.0} & \textbf{94.8} & \textbf{96.4} \\
                                   & Ours                    & \textbf{66.1}    & \textbf{81.2} & \textbf{55.4} & \textbf{34.7}    & \textbf{91.2}            & 98.2             & \textbf{96.7}    & \underline{92.8}          & \underline{95.9}          \\
        \bottomrule 
    \end{tabular}
    % \vspace{-5mm}
\end{table*}

\begin{table}[ht]
    \footnotesize
    \caption{The mean rank-1 accuracy (\%) on OUMVLP excluding the under different skeleton representations, excluding identical-view cases.}
    \label{tab2}
    % \vspace{1mm}
    \centering
    \begin{tabular}{c|c|c}
        \toprule
        Skeleton Input           & Method                & Rank-1 \\
        \midrule
        \multirow{2}{*}{Point}   & BiFusion\cite{9-BiFusion}              & 89.9   \\
                                 & MMGaitFormer\cite{10-MMGaitFormer}          & 90.1   \\
        \midrule
        Joint-based heatmap      & \multirow{2}{*}{Ours} & \underline{90.8} \\
        Joint/Limb-based heatmap &                       & \textbf{91.2}    \\
        \bottomrule    
    \end{tabular}
    % \vspace{-5mm}
\end{table}
    
\subsection{Loss Function}
    To achieve optimal performance, we employ triplet loss \cite{20-triplet-loss}, cross-entropy loss, and Wasserstein loss \cite{21-wasserstein-loss} to train GaitMA. 
    
    First, the network is trained to converge by optimizing the classification space using cross-entropy loss which can be formulated as:
    \begin{eqnarray}
        {\cal L}_{\mathrm{ce}}&=&-{\frac{1}{N}}\sum_{i=1}^{N}log{\frac{e^{W_{y_{i}}^{T}x_{i}+b_{y_{i}}}}{\sum_{j=1}^{n}e^{W_{j}^{T}x_{i}+b_{j}}}}, 
    \end{eqnarray}
    where $x_{i}$ is the feature of the $i$-th sample, and its label is $y_{i}$.
    
    Second, triplet loss is proposed to enable the model to find a more discriminative metric space by optimizing distances, which can be defined as:
    \begin{eqnarray}
    {\cal L}_{tri}&=&\varphi \left [ D\left ( F_{i},F_{k}  \right ) - D\left ( F_{i},F_{j} \right ) +m \right ] . 
    \end{eqnarray}
    $\varphi$ is equal to $max\left ( \alpha, 0 \right )$, $D\left ( F_{i}, F_{k}  \right )$ represents the Euclidean distance between the features of sample i and sample k, m denotes the margin for the triplet loss.
    
    Finally, we introduce the Wasserstein loss to minimize the distance between the two modalities, ensuring effective fusion and accelerating the convergence of the model. Assuming that the identity features follow a normal distribution, we can utilize online estimations to calculate the means and covariance matrices of the identity features:
    \begin{eqnarray}
        \tilde{Y_{1}}\sim{\mathcal{N}}(\mu,\Sigma),\tilde{Y_{2}}\sim\mathcal{N}(\mu^{*},\Sigma^{*}).\label{6}
    \end{eqnarray} 
    The similarity between these two Gaussian distributions is measured using the 2-Wasserstein distance, which results in the Wasserstein loss:
    \begin{eqnarray}
        {\cal L}_{w}\ {\stackrel{\triangle}{=}}\ W_{2}(\tilde{Y_{1}},\tilde{Y_{2}})=||\mu-\mu^{*}||_{2}^{2}+||\Sigma^{\frac{1}{2}}-\Sigma^{*}{}^{\frac{1}{2}}||_{F}^{2}.\label{7}
    \end{eqnarray}
    
    The joint loss function can be expressed as follows:
    \begin{eqnarray}
        {\cal L}&=&{\alpha_1}{\cal L}_{\mathrm{tri}}+{\alpha_2}{\cal L}_{\mathrm{ce}}+{\alpha_3}{\cal L}_{\mathrm{w}},\label{5}
    \end{eqnarray} 
    where the hyper-parameters ${\alpha_1}$, ${\alpha_2}$ and ${\alpha_3}$ are balance factors to weight the losses to each other, where ${\alpha_1}$ = 1.0, ${\alpha_2}$ = 0.1 and ${\alpha_3}$ = 0.1 respectively.

\section{Experiments}

\subsection{Datasets}
     We evaluated our proposed method on three commonly used datasets, including one outdoor dataset: Gait3D \cite{23-Gait3D} and two indoor datasets: CASIA-B \cite{24-CASIAB}, OU-MVLP \cite{25-OUMVLP}.
    
    \textbf{Gait3D} \cite{23-Gait3D} is a large-scale gait dataset captured in the wild, comprising 4,000 subjects and 25,309 sequences. The dataset features 25,309 sequences acquired through camera capture and provides four modalities: silhouettes, 2D and 3D coordinates of joints, and 3D meshes. It is divided into a training set containing 3,000 subjects and a test set consisting of 1,000 subjects.

    \textbf{OU-MVLP} \cite{25-OUMVLP} contains 10307 subjects, and each subject includes 28 sequences obtained from 14 camera views. For each view, each subject has 2 sequences (NM\#01 and NM\#02). The sequences of the first 5153 subjects were used for training, and the sequences of the remaining 5154 subjects were used for testing.

    \textbf{CASIA-B} \cite{24-CASIAB} is one of the earliest widely used gait datasets, consisting of 124 subjects. Each subject is represented with 11 views, and each view contains ten sequences. These sequences are captured under three different walking conditions: normal walking (NM), walking with a bag (BG), and walking in a coat (CL). The dataset is divided into two parts: the first 74 subjects are designated as the training set, while the remaining 50 subjects constitute the test set.

\subsection{Experimental Settings}
    For CASIA-B and OU-MVLP, the resolution of silhouettes we take is $64 \times 44$. For Gait3D, the resolution of silhouettes we take is $128 \times 88$. We use SGD as the optimizer for the training model in both CASIA-B, OU-MVLP, and Gait3D. The initial learning rate and the weight decay of the SGD optimizer as 0.1 and 0.0005. 
    For CASIA-B, we train our model for 60k with (8, 16) batch size, the learning rate is set to 1e-2 at the 20k iteration and 1e-3 at the 40k iteration respectively. For OU-MVLP, the total iteration is 150k with (32, 8) batch size, decaying the learning rate to 1e-2 and 1e-3 at the 50k and 100k iterations.
    For Gait3D, the batch size is set to (16, 4), the total iteration is 60k.
    % the total iteration is 60k. 
    % and the learning rate is reduced to 1e-2, 1e-3, 1e-4 at 20k, 40k, and 50k. 
    
\subsection{Comparison with State-of-the-art Methods}
    We compare GaitMA to other state-of-the-art (SOTA) gait recognition work, with comparative results detailed in Table~\ref{tab1}. This comparison encompasses methods based on silhouette-based, skeleton-based, and multimodal three mainstream approaches. Additionally, a focused comparison with other multimodal methods, specifically in terms of skeleton representation, is presented in Table~\ref{tab2}. These comprehensive evaluation results are sourced from the respective original publications.

    \textbf{Comparison with silhouette-based methods:} GaitMA exhibits superior performance on the CASIA-B, OU-MVLP, and Gait3D datasets, underscoring the enhanced gait characterization achieved through the integration of the skeleton feature. This is particularly evident in challenging scenarios, such as the real-world Gait3D dataset and the CASIA-B(CL) dataset, where silhouette quality is compromised by complex backgrounds, occlusions, and camera angles. The incorporation of skeleton features in our method not only demonstrates significant improvements in these conditions but also provides an effective resolution to these challenges. Notably, our method outperforms the current leading GaitBase method by a margin of 1.5\%.

    \textbf{Comparison with skeleton-based methods:} Our approach surpasses current skeleton-based methods, which are hindered by the limited accuracy of pose estimation algorithms and the absence of spatial shape features, rendering them less competitive, particularly on large-scale and real-world datasets. Notably, our method achieves a 43.7\% higher Rank-1 accuracy on Gait3D compared to GPGait.
    
    \textbf{Comparison with multi-model methods:} Diverging from prevalent multimodal approaches that utilize coordinate points, our method transforms joint points into joint/limb-based heatmaps, enhancing skeleton feature representation. We present a comparison of this method with point-skeleton input methods and different heatmap forms on the OU-MVLP dataset in Table~\ref{tab2}. Table~\ref{tab1} outlines our evaluation across the full datasets. Here, we initially apply our multimodal strategy to the real-world Gait3D dataset, subsequently achieving state-of-the-art results on the large-scale OU-MVLP dataset. The CASIA-B dataset, comprising 124 individuals in a simplistic indoor setting, presents a risk of overfitting in large models, which may degrade generalization in test scenarios, we believe that CASIA-B is no longer suitable as a benchmark dataset. Notably, our approach registers an improvement of 1.3\% and 1.1\% on the OU-MVLP dataset, surpassing BiFusion and MMGaitFormer, respectively.
    
\begin{table}[t]
    \vspace{-0.9cm}
    \footnotesize
    \renewcommand\arraystretch{0.8}
    \caption{Ablation study on the effectiveness of each individual module on the Gait3D}
    \label{tab3}
    % \vspace{1mm}
    \centering
    \begin{tabular}{l@{\hspace{2mm}}lclll@{\hspace{2mm}}c@{\hspace{2mm}}c}
        \toprule
        Methods                               & Sil        & ${\cal J}$ and ${\cal L}$ & CAM        & MLM        & ${\cal L}_{\mathrm{w}}$ & Rank-1 & mAP \\
        \midrule
        Baseline                              & \checkmark &                           &            &            &                         & 59.9  & 48.9 \\
        + ${\cal J}$ and ${\cal L}$           & \checkmark & \checkmark                &            &            &                         & 64.1  & 52.8 \\
        + CAM                                 & \checkmark & \checkmark                & \checkmark &            &                         & 64.5  & 53.2  \\
        + MLM                                 & \checkmark & \checkmark                & \checkmark & \checkmark &                         & 65.3  & 54.2  \\
        + ${\cal L}_{\mathrm{w}}$             & \checkmark & \checkmark                & \checkmark & \checkmark & \checkmark              & 66.1  & 55.4 \\
        \bottomrule
    \end{tabular}
    % \vspace{-5mm}
\end{table}

\subsection{Ablation Study}

    To validate the efficacy of each component in GaitMA, including joint/limb-based heatmaps, which provides robust skeleton gait features, CAM and MLM for spatial and temporal multi-model feature fusion, Wasserstein loss which makes the distribution of fused features as similar as possible, we conduct ablation studies the Gait3D dataset with results in Table~\ref{tab3}. Furthermore, we demostrate the universality of our method by applying it to two state-of-the-art silhouette-based gait recognition models, the evaluation results of which are displayed in Table~\ref{tab4}.

    \textbf{Ablation Study of joint/limb-based heatmaps.} To investigate the impact of incorporating the skeleton branch, which is represented by joint/limb-based heatmaps, we devise a baseline model consisting solely of a single silhouette branch. Remarkably, the inclusion of skeletons leads to a substantial increase in accuracy by \textbf{3.8\%}, thus performing a significant improvement in the gait recognition task.

    \textbf{Ablation Study of CAM\&MLM.} 
    The integration of these two modules improves the accuracy by \textbf{1.2\%} compared to a simple element-wise addition approach. 
    Specifically, CAM yields a \textbf{0.4\%} improvement, while MLM achieves a \textbf{0.8\%} improvement, demonstrating the effectiveness of each module. 
    The results highlight that the two modules we proposed effectively facilitate the fusion of two modalities, resulting in a more comprehensive and robust gait representation.

    \textbf{Ablation Study of Wasserstein loss.}
    Wasserstein loss makes the distribution of fused features as similar as possible for each identity.
    When training GaitMA using Wasserstein loss, the accuracy improved by \textbf{0.8\%}, demonstrating that the introduction of Wasserstein loss ensures effective fusion and accelerates the convergence of the model.

    \textbf{Universality of GaitMA.}
    we demonstrate the universality of our method by applying it to two state-of-the-art silhouette-based gait recognition models, i.e., GaitSet\cite{1-gaitset}, GaitPart\cite{2-gaitpart}. We denote the models after applying our method as GaitSet-MA and GaitPart-MA. The integrated model incorporates the original structures of GatiSet and GaitPart to encode silhouette features. Then, we introduce joint/limb-based skeleton feature encoding to extract spatial shape information from the skeleton modality and introduce CAM and MLM to realize the fusion of multimodal feature information.

    The results on the Gait3D datasets, as presented in Table~\ref{tab4}, demonstrate the effectiveness of our proposed method. It exhibits significant improvements in Rank-1 accuracy, with an increase of 36.7 to 48.2 for GaitSet and 28.5 to 45.8 for GaitPart. This consistent enhancement is observed across both models, highlighting the efficacy of our approach.

\begin{table}[t]
    \vspace{-0.9cm}
    \footnotesize
    \renewcommand\arraystretch{0.8}
    \centering
    \caption{Universality study results on the Gait3D dataset}
    \label{tab4}
    % \vspace{1mm}
    \begin{tabular}{c|c|c}
        \toprule
        Method    & Modality                                                       & Rank-1 \\
        \midrule
        GaitSet\cite{1-gaitset}   & Sihouette                                                      & 36.7       \\
        GaitSet-MA  & Silhouette+Skeleton                                            & 48.2       \\
        \midrule
        GaitPart\cite{2-gaitpart}  & Silhouette                                                     & 28.2       \\
        GaitPart-MA & Silhouette+Skeleton                                            & 45.8       \\
        \midrule
        Ours      & Silhouette+Skeleton                                            & 66.1       \\
        \bottomrule
    \end{tabular}
    % \vspace{-5mm}
\end{table}

\section{Conclusion}
% In this paper, we proposed a multi-model fusion framework for gait recognition, namely GaitMA which significantly improves the accuracy of gait recognition by effectively fusing two modalities. Adequate experiments prove the effectiveness of our framework and illustrate that multi-modal fusion has a prospect in gait recognition.

 % This paper introduces DiffGait, the first successful method to generate dense body shapes from skeletal representations which contains advanced knowledge of human structure. Then, We propose a novel skeleton-based framework (DGR-Net) that achieves state-of-the-art cross-domain performance and the most consistent within-domain performance on four mainstream datasets, both indoor and outdoor, in comparison to other skeleton-based methods. Specifically, the use of Heat-skeletons to represent bones provides more explicit human structural features and demonstrates greater robustness in real scenarios, while we use DiffGait to recover the dense body shape Diff-Silhouette from them to provide more highly distinguishable gait features. Furthermore, our well-designed Perceptual Gait Fusion Module can effectively combine two modalities to obtain a more robust and comprehensive gait representation for recognition. We aim to delve deeper into the application of diffusion modeling to gait representation interaction in the future, aiming to further advance the development of gait recognition.

This paper introduces GaitMA, a novel multi-modal gait recognition framework that effectively combines two modalities to obtain a more robust and comprehensive gait representation for recognition. Compared to other multi-modal gait recognition approaches, our method consistently demonstrates superior performance across three mainstream datasets, both indoor and outdoor, and marks the first application of multi-modal methods in the wild. Specifically, the use of Heat-skeletons representations provides clearer structural features of the human body and exhibits enhanced robustness in real scenarios. Furthermore, our well-designed Co-attention alignment module and Mutual learning module, along with the introduction of Wasserstein loss, effectively eliminate redundant features between modalities and integrate efficient gait representations. Our goal is to continue advancing the study of multi-modal feature learning within the field of gait recognition, thereby continuously propelling progress in gait recognition.

\section*{Acknowledgment}
    This work is supported in part by the National Natural Science Foundation of China (Grant No. 42106193, 41927805).
    
\newpage
\bibliography{refs}





%\bibliographystyle{IEEEtran}

\end{document}